\pdfoutput=1

\documentclass[11pt]{article}

\usepackage{ACL2023}

\usepackage{times}
\usepackage{latexsym}
\usepackage{booktabs} 
\usepackage{tabularx}
\usepackage{listings}
\usepackage{xcolor}
\usepackage{graphicx}

\lstdefinestyle{jsonstyle}{
  basicstyle=\small\ttfamily,
  commentstyle=\color{gray},
  keywordstyle=\color{blue},
  numberstyle=\tiny\color{purple},
  stringstyle=\color{green},
  morestring=[b]",
  morestring=[b]',
  morecomment=[l]{//},
  morecomment=[s]{/*}{*/},
  frame=single,
  framesep=5pt,
  aboveskip=10pt,
  belowskip=10pt,
  breaklines=true,
  showstringspaces=false,
  captionpos=b,
  numbers=left,
  numbersep=5pt,
  escapechar=|
}
\usepackage[T1]{fontenc}

\usepackage[utf8]{inputenc}

\usepackage{microtype}

\usepackage{inconsolata}
\usepackage{textcomp}
\usepackage[T1]{fontenc}
\usepackage[utf8]{inputenc}
\usepackage{microtype}
\usepackage{inconsolata}
\usepackage{textcomp}

\title{Legend at ArAIEval Shared Task: Persuasion Technique Detection using a Language-Agnostic Text Representation Model}

\author{
  Olumide E. Ojo\textsuperscript{1,5,a}, Olaronke O. Adebanji\textsuperscript{1,b}, Hiram Calvo\textsuperscript{1,c}, Damian O. Dieke\textsuperscript{3,d}, \\
  Olumuyiwa E. Ojo\textsuperscript{4,e}, Seye E. Akinsanya\textsuperscript{2,f}, Tolulope O. Abiola\textsuperscript{2,g}, Anna Feldman\textsuperscript{5,h}\\
  \small\textsuperscript{1}Instituto Politécnico Nacional, Centro de Investigación en Computación, CDMX, Mexico\\ \small\textsuperscript{2}Federal University Oye-Ekiti, Nigeria; \textsuperscript{3}Caritas University, Nigeria; \textsuperscript{4}Lead City University, Nigeria; \textsuperscript{5}Montclair State University, USA \\
 \small{\{\textsuperscript{a}olumideoea, \textsuperscript{b}olaronke.oluwayemisi, \textsuperscript{c}hiramcalvo, \textsuperscript{d}dieketobesky, \textsuperscript{e}muyeskin, \textsuperscript{f}akinsanyaseye, \textsuperscript{g}abiolato92\}@gmail.com};\\
  \small{\textsuperscript{h}feldmana@montclair.edu}
}

\begin{document}
\maketitle
\begin{abstract}
In this paper, we share our best performing submission to the Arabic AI Tasks Evaluation Challenge (ArAIEval) at ArabicNLP 2023. Our focus was on Task 1, which involves identifying persuasion techniques in excerpts from tweets and news articles. The persuasion technique in Arabic texts was detected using a training loop with XLM-RoBERTa, a language-agnostic text representation model. This approach proved to be potent, leveraging fine-tuning of a multilingual language model. In our evaluation of the test set, we achieved a micro F1 score of 0.64 for subtask A of the competition.

\end{abstract}

\section{Introduction}
In an era defined by the rapid dissemination of information through digital channels, the task of recognizing persuasion techniques in text is now more crucial than ever~\citep{araieval:arabicnlp2023-overview, hossain2021csecu, gupta2021volta, sadeghi2023sinaai, alam-etal-2022-overview, abujaber2021lecun}. The advent of the Internet and social media has created new avenues for influence and manipulation~\citep{dholakia2023miasma, ruffo2023studying, botes2023autonomy}. Although these technological advances have undoubtedly given people unparalleled access to information and a platform to express their thoughts, they have also introduced new avenues for persuasion, influence, and even propaganda. At the heart of this shift in our way of life is the fundamental challenge of distinguishing between genuinely informative and impartial content and content subtly crafted to promote a specific agenda or ideology. A critical component of the fight against spreading misinformation is the development of tools and resources in NLP that can detect persuasion technique in news articles and posts on social media.

Arabic language is among the most spoken languages in the world~\citep{ghazzawi1992arabic}. The Arabic speaking world stands out in importance due to its intricate mix of language, culture, and geography. The diversity of linguistic expressions in Arabic extends beyond the spoken word and permeates every aspect of life, including digital text. Arabic connects a vast and diverse population of native speakers and foreigners with its rich heritage and multiple dialects. Its influence extends over a wide territory, from Arab nations in the southern part of the Arabian Peninsula, to Asia, and to the Maghreb in North Africa and the heart of the Arabian Peninsula~\citep{huafeng2019history}. In this digital age, the importance and extensive use of Arabic in various forms reflects the profound impact of technology on this linguistic community. It has provided a platform for Arabic speakers around the world to engage in dialogue, share ideas, and express their thoughts in a global context. The diverse Arab-speaking populations foster a rich and dynamic environment where individuals can connect, collaborate, and debate issues of global significance. However, the proliferation of digital media in Arabic also presents challenges. Digital media have become a fertile ground for the dissemination of persuasive content, including propaganda, misinformation, and various forms of manipulation~\citep{aleroud2023span, abd2023hybrid}. The use of technology in preserving the integrity of the Arabic language and ensuring responsible use of digital media is of utmost importance. 

In today's world, we are surrounded by information, especially on the Internet and social media. Text classification can serve as a foundational step for the detection of the persuasion technique in text on social media and the Internet. These texts can be classified according to their emotional tone using sentiment analysis techniques~\citep{nikolaidis2023experiments, ojo2022automatic, ojo2021performance, ojo2020sentiment,  ojo2023transformer, ojo2022language, piskorski2023semeval, hromadka2023kinitveraai}. Persuasion techniques can be tricky to spot because they come in many forms, such as stories, logical arguments, or even subtle language tricks to sway our thinking. These techniques are powerful and are not always used for good reasons. To deal with this, researchers and experts are working on ways to detect when someone is trying to persuade people through text. In this way, we will be able to recognize when we are influenced and when people are spreading false or misleading information, which can be harmful. In this paper, we develop a methodology to automatically identify and analyze persuasion techniques in text using the XLM-RoBERTa model. Using the power of NLP, we can harness the capabilities of state-of-the-art models and unravel the persuasion techniques embedded within Arabic text, contributing to both media ethics and the combating of misinformation. The intricacies of persuasion technique detection in Arabic are discussed in detail, along with the possibility of applying this knowledge across different languages.

\section{Related Work}
A significant amount of research has been conducted on the detection of persuasion techniques in text~\citep{araieval:arabicnlp2023-overview, modzelewski2023dshacker, hossain2021csecu, sadeghi2023sinaai, abujaber2021lecun}. Researchers have explored binary and multilabel approaches to detecting persuasion techniques.

In their article, the authors in~\citep{modzelewski2023dshacker} focused on detecting genres and persuasion techniques in multiple languages using various data augmentation techniques to enhance their models. For genre detection, they created synthetic texts using the GPT-3 Davinci language model, while for persuasion technique detection, they augmented the dataset using text translation with the DeepL translator. Their fine-tuned models achieved top ten rankings in all languages, demonstrating the effectiveness of their approach. They also excelled in genre detection, securing top positions in Spanish, German, and Italian. They presented a single multilingual system using the RoBERTa model to classify online news genres and improved this system by adding texts generated by the GPT3-Davinci model to the training dataset.

\citep{hasanain2023qcri} addressed misinformation in mainstream and social media and the challenges faced by manual detection and verification efforts by journalists and fact checkers in the SemEval-2023 task~\cite{piskorski2023semeval}. The task included three subtasks, six languages, and three surprise test languages, totaling 27 different test scenarios. The authors successfully submitted entries for all 27 test setups and the official results placed their system among the top three for 10 of these setups. They fine-tuned transformer models in the multiclass and multilabel classification settings, experimenting with both monolingual and multilingual pre-trained models, as well as data augmentation. Their multilingual model based on XLM-RoBERTa demonstrated superior performance in all tasks, even for languages not seen during training.

The most effective solution in the detection of persuasion techniques for Subtask 3 of SemEval 2023 Task 3 by~\citep{hromadka2023kinitveraai} delivered promising performance, with micro-F1 scores ranging from 36 to 55\% for languages seen during training and 26 to 45\% for languages unseen. Given the multilingual nature of the data and the presence of 23 labels (resulting in limited labeled data for some language-label combinations), the authors chose to fine-tune pre-trained transformer-based language models. Through extensive experimentation, they identified the optimal configuration, featuring a large multilingual model (XLM-RoBERTa) trained on all input data, with carefully calibrated confidence thresholds for known and surprise languages separately. Their final system demonstrated superior performance, ranking first in six of nine languages, including two surprise languages, and achieving highly competitive results in the remaining three languages. 

The experiments of~\citep{nikolaidis2023experiments} focused on the detection of persuasion techniques in online news articles in Polish and Russian. These experiments used a taxonomy comprising 23 distinct persuasion techniques. Persuasion techniques were evaluated in several ways, including the granularity of the classification (coarse with six labels or fine with 23 labels) and the level of location of the labels (subword, sentence, paragraph). The study compared the performance of state-of-the-art transformer-based models trained both monolingually and multilingually. The findings indicate that multilingual models generally outperform monolingual models in various evaluation scenarios. However, due to the complexity of the task, there remains substantial room for improvement in the field of persuasion technique detection within online news articles.

Inspired by~\citep{hasanain2023qcri, hromadka2023kinitveraai}, our research focus is to determine whether a multi-genre snippet (tweets and news paragraphs of news articles) contains a persuasion technique or not. This is a binary classification task, and we are categorizing Arabic text either as containing a persuasion technique or as lacking persuasion technique. Our approach involves fine-tuning a large pre-trained language model based on transformers. We conducted experiments with different language models and concluded with XLM-RoBERTa due to its superior performance. Furthermore, we fine-tuned our system by tweaking hyperparameters and conducting multiple iterations. This process involved calculating cross-entropy loss, performing backpropagation, and updating the model's weights.

\section{Persuation Technique Detection}
\subsection{Dataset Analysis}
Datasets provided by the organizers consist of a diverse collection of Arabic text samples, each associated with a label. These texts have been carefully selected to represent where persuasion techniques are present or not. The dataset were labeled as 'true' or 'false', where the true class represents texts where persuasion techniques are used to influence the reader's opinion, and the false class consists of texts that do not use persuasion techniques. The method involves binary classification to determine the presence of a persuasion technique within the document.

\begin{figure}[ht]
    \centering
    \includegraphics[width=\linewidth]{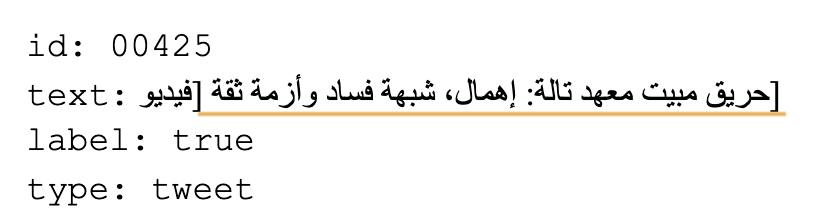}
    \caption{Persuation technique in the Arabic text}
    \label{fig:arabic}
\end{figure}

The dataset comprises 2,427 samples in the training set and 503 in the test set. Within the training data, there is an imbalance in the label distribution, with 1,918 samples labeled as "true" and 509 labeled as "false". An example of text that represents the persuasion technique in the dataset is shown in Figure~\ref{fig:arabic}.

\subsection{Application of XLM-RoBERTa}
Leveraging the Hugging Face Transformers library, we prepared and formatted the data so that it could be seamlessly incorporated into our model. In line with the methodology outlined in Section 4, we designed a custom model architecture and then proceeded to encode and load the data for further processing. During training, we optimized the model's classification head to minimize cross-entropy loss, and predictions are made on unseen text, followed by post-processing. This approach demonstrates the adaptability and effectiveness of XLM-RoBERTa in diverse linguistic contexts, offering practical solutions to automate the detection of persuasion techniques across languages.

\section{System Setup and Experiments}
\subsection{Training Strategy}
Our training strategy involves fine-tuning the XLM-RoBERTa model using the provided dataset. To optimize the performance of the model, we incorporate several key components and hyperparameters, which are summarized in Table \ref{table:hyperparameters}.

\begin{table*}[ht]
\centering
\begin{tabular}{lr}
\toprule
\textbf{Hyperparameter} & \textbf{Value} \\
\midrule
Learning rate & $5 \times 10^{-5}$ \\
Batch size & 16 \\
Epochs & 6 \\
Optimizer & ADAM \\
Early stopping & Cross-entropy loss \\
Learning rate scheduler & StepLR (factor=0.85, step size=2) \\
\bottomrule
\end{tabular}
\caption{Model Hyperparameters}
\label{table:hyperparameters}
\end{table*}

\subsection{Model Fine-Tuning}
During model fine-tuning, we added a classification layer at the end of the pre-trained XLM-RoBERTa model. This additional layer allows the model to perform the specific classification task required. To prevent overfitting, a dropout layer was incorporated.

\subsection{Class Weights}
To address the class imbalance in the dataset, we adjusted the learning process using class weights. This ensures that the model effectively learns from all classes and is not biased by data imbalance.

\subsection{Evaluation Metric}
The model's performance was evaluated using the micro-score F1, the default metric for this subtask. This metric provides a comprehensive measure of the model's classification performance.

\subsection{Training Process}
The training process involved conducting a total of 6 epochs, where each epoch represents a complete pass through the entire training dataset. To avoid overfitting, early stopping was employed on the basis of cross-entropy loss.

\subsection{Learning Rate Scheduler}
A learning rate scheduler was implemented to dynamically adjust the learning rate during training. Specifically, we used a StepLR scheduler with a reduction factor of 0.85 applied every 2 epochs. This scheduling strategy contributes to training stability and controlled convergence.

\subsection{Optimization Process}
To initiate the optimization process, we reset the gradients (setting them to zero). Subsequently, we conducted backpropagation to calculate gradients for all model parameters and, lastly, update the model's parameters using the computed gradients.

Our approach leverages these components and hyperparameters to fine-tune the model effectively, ensuring robust and controlled training.

\section{Results}
The evolution of pre-trained language models has ushered in significant advancements across different NLP tasks. In this section, we present the results of our experiments on persuasion technique detection in Arabic text. Our findings revealed the versatility of XLM-RoBERTa in effectively handling multilingual data. We evaluated the model using the micro-F1 score and an overview of the results achieved by our model on the dataset is shown in Table~\ref{tab:results}.

\begin{table}[h]
\centering
\begin{tabular}{@{}lcc@{}} 
\toprule
\textbf{Model} & \textbf{F1-Score} \\
\midrule
Baseline model & 0.4771 \\
XLM-RoBERTa & 0.6402 \\
\bottomrule
\end{tabular}
\caption{Performance Comparison of Models for Persuasion Technique Detection in Arabic Text}
\label{tab:results}
\end{table}

From Table~\ref{table:hyperparameters}, our findings reveal the effectiveness of XLM-RoBERTa in identifying persuasion techniques within the Arabic text.  The model's pre-trained knowledge, combined with its cross-lingual capabilities, makes it a promising tool for similar tasks in other languages as well. As Arabic is a morphologically rich language with various dialects, the success of XLM-RoBERTa in this task is a testament to its robustness and versatility.

\section{Conclusion}
The application of XLM-RoBERTa, a language-agnostic text representation model,  for the detection of the persuasion technique in Arabic text illustrates the growing potential of cross-lingual models in specialized NLP tasks.  Arabic, with its rich morphology and diverse dialects, presents unique challenges for text analysis. Our proposed model has the ability to capture the underlying structure and semantics of persuasion technique in text, regardless of language. The results obtained in our analysis demonstrate that XLM-RoBERTa can adapt effectively and perform well on such intricate tasks, even in languages that are structurally different from the ones they were originally trained on. This not only underscores the versatility of XLM-RoBERTa but also sets a promising direction for further research in detecting persuasion techniques across various languages. In future work, we plan to accommodate more languages in the dataset, and fine-tune other multilingual models for this task.

\section*{Ethics Statement}
We acknowledge the influence that scientific research can have on society and recognize our responsibility to ensure its positive impact. Our research is carried out with the aim of addressing real-world challenges, promoting well-being, and improving quality of life. We actively seek feedback and input from those affected by our research. In the event that our research may have unintended negative impacts, we are committed to addressing and rectifying those issues promptly.

\section*{Acknowledgments}
This work was done with partial support from the Mexican Government through the grant A1-S-47854 of CONACYT, Mexico, grants 20232138, 20230140, 20232080 and 20231567 of the Secretaría de Investigación y Posgrado of the Instituto Politécnico Nacional, Mexico. The authors thank CONACYT for the computing resources brought to them through the Plataforma de Aprendizaje Profundo para Tecnologías del Lenguaje of the Laboratorio de Supercómputo of the INAOE, Mexico and acknowledge the support of Microsoft through the Microsoft Latin America PhD Award.

\bibliography{custom}
\bibliographystyle{acl_natbib}
\end{document}